\documentclass[conference]{IEEEtran}
\IEEEoverridecommandlockouts
\usepackage{cite}
\usepackage{xurl}
\usepackage{amsmath,amssymb,amsfonts}
 \usepackage[titlenumbered,ruled,noend]{algorithm2e}
\usepackage{tabularx}
\usepackage{graphicx}
\usepackage{textcomp}
\usepackage[driverfallback=dvipdfm]{hyperref}
\usepackage{xcolor}
\usepackage{booktabs}
\usepackage{multirow}
\usepackage{enumitem}   
\usepackage{cleveref}
\usepackage{lipsum}
\usepackage{listings}
\usepackage{stfloats}
\usepackage{float}
\usepackage[bottom]{footmisc}
\fnbelowfloat

\def\BibTeX{{\rm B\kern-.05em{\sc i\kern-.025em b}\kern-.08em
T\kern-.1667em\lower.7ex\hbox{E}\kern-.125emX}}
\begin{document}

\lstset{
    language=HTML,
    basicstyle=\scriptsize,
    breaklines=true
}

\makeatletter % changes the catcode of @ to 11
\newcommand{\linebreakand}{%
  \end{@IEEEauthorhalign}
  \hfill\mbox{}\par
  \mbox{}\hfill\begin{@IEEEauthorhalign}
}
\makeatother % changes the catcode of @ back to 12

\title{Dark patterns in e-commerce: a dataset and its baseline evaluations}

% https://ieeexplore-ieee-org.waseda.idm.oclc.org/stamp/stamp.jsp?tp=&arnumber=9556241&tag=1
\author{
    \IEEEauthorblockN{
        Yuki Yada\IEEEauthorrefmark{1}, Jiaying Feng\IEEEauthorrefmark{1}, Tsuneo Matsumoto\IEEEauthorrefmark{2}, Nao Fukushima\IEEEauthorrefmark{3}, Fuyuko Kido\IEEEauthorrefmark{4}\IEEEauthorrefmark{5},  Hayato Yamana\IEEEauthorrefmark{1}
    }
    \IEEEauthorblockA{\IEEEauthorrefmark{1} \textit{Waseda University}, Tokyo, Japan, E-mail: \{yada\_yuki, kayouh, yamana\}@yama.info.waseda.ac.jp}
    \IEEEauthorblockA{\IEEEauthorrefmark{2} \textit{National Consumer Affairs Center of Japan}, Sagamihara, Kanagawa, Japan, Email: tsuneo.matsumoto@aoni.waseda.jp}
    \IEEEauthorblockA{\IEEEauthorrefmark{3} \textit{LINE Corporation}, Shinjuku, Tokyo, Japan, Email: nao.fukushima@linecorp.com}
    \IEEEauthorblockA{\IEEEauthorrefmark{4} \textit{Waseda Research Institute for Science and Engineering}, Tokyo, Japan, Email: fkido@aoni.waseda.jp}
    \IEEEauthorblockA{\IEEEauthorrefmark{5} \textit{National Institute of Informatics}, Chiyoda-ku, Tokyo, Japan}
}
% \author{

% \IEEEauthorblockN{Yuki YADA}
% \IEEEauthorblockA{\textit{Dept. of CSCE, Waseda University}\\
% Tokyo, Japan \\
% yada\_yuki@yama.info.waseda.ac.jp}
% \and 
% \IEEEauthorblockN{Jiaying FENG}
% \IEEEauthorblockA{
% \textit{Dept. of CSCE, Waseda University}\\
% Tokyo, Japan \\
% kayouh@yama.info.waseda.ac.jp}
% \and
% \IEEEauthorblockN{Tsuneo MATSUMOTO}
% \IEEEauthorblockA{
% \textit{National Consumer Affairs Center of Japan} \\
% Sagamihara, Kanagawa, Japan \\
% tsuneo.matsumoto@aoni.waseda.jp}
% \and

% \linebreakand % <----- NOTE HERE, breaking after the third one!

% \IEEEauthorblockN{Nao FUKUSHIMA}
% \IEEEauthorblockA{
% \textit{LINE Corporation} \\
% Shinjuku, Tokyo, Japan \\
% nao.fukushima@linecorp.com}
% \and
% \IEEEauthorblockN{Fuyuko KIDO}
% \IEEEauthorblockA{
% \textit{Waseda Research Institute for Science and Engineering} \\
% Tokyo, Japan \\
% \textit{National Institute of Informatics,} \\ 
% Chiyoda-ku, Tokyo, Japan \\
% fkido@aoni.waseda.jp}
% \and
% \IEEEauthorblockN{Hayato YAMANA}
% \IEEEauthorblockA{
% \textit{Faculty of Sci. and Eng.,} \\ 
% \textit{Waseda University} \\
% Tokyo, Japan \\
% yamana@yama.info.waseda.ac.jp}
% }

\maketitle

\begin{abstract}
 Dark patterns, which are user interface designs in online services, induce users to take unintended actions. Recently, dark patterns have been raised as an issue of privacy and fairness. Thus, a wide range of research on detecting dark patterns is eagerly awaited. In this work, we constructed a dataset for dark pattern detection and prepared its baseline detection performance
with state-of-the-art machine learning methods. The original dataset was obtained from Mathur et al.’s study in 2019 \cite{11kScale}, which consists of 1,818 dark pattern texts from shopping sites. Then, we added negative samples, i.e., non-dark pattern texts, by retrieving texts from the same websites as Mathur et al.’s dataset. We also applied state-of-the-art machine learning methods to show the automatic detection accuracy as baselines, including BERT, RoBERTa, ALBERT, and XLNet. As a result of 5-fold cross-validation, we achieved the highest accuracy of 0.975 with RoBERTa. The dataset and baseline source codes are available at \url{https://github.com/yamanalab/ec-darkpattern}.
\end{abstract}


\section{Introduction}\label{sec:introduction}
\subsection{Dark Patterns}
Dark patterns are user interface designs on online services that make users behave in unintended ways. Dark patterns have been called into question in recent years.

In 2010, Harry \cite{darkpattern.org} defined dark patterns as “tricks used in websites and apps that make a user do things that the user did not mean to, like buying or signing up for something.” Fig. \ref{dpex} shows an example of dark patterns, classified as Obstruction \cite{11kScale}. The obstruction makes it difficult for users to conduct a specific task, such as cancellation and withdrawal. For example, in Fig. \ref{dpex}, the website only shows an “ADD TO CART” button but no cancel button. Instead, the website shows “If you would like to cancel your membership, please call ... or contact us via email at ...”, which makes it difficult for the user to cancel by limiting the cancellation to telephone calls or email.

\begin{figure*}[h]
\includegraphics[width=\linewidth]{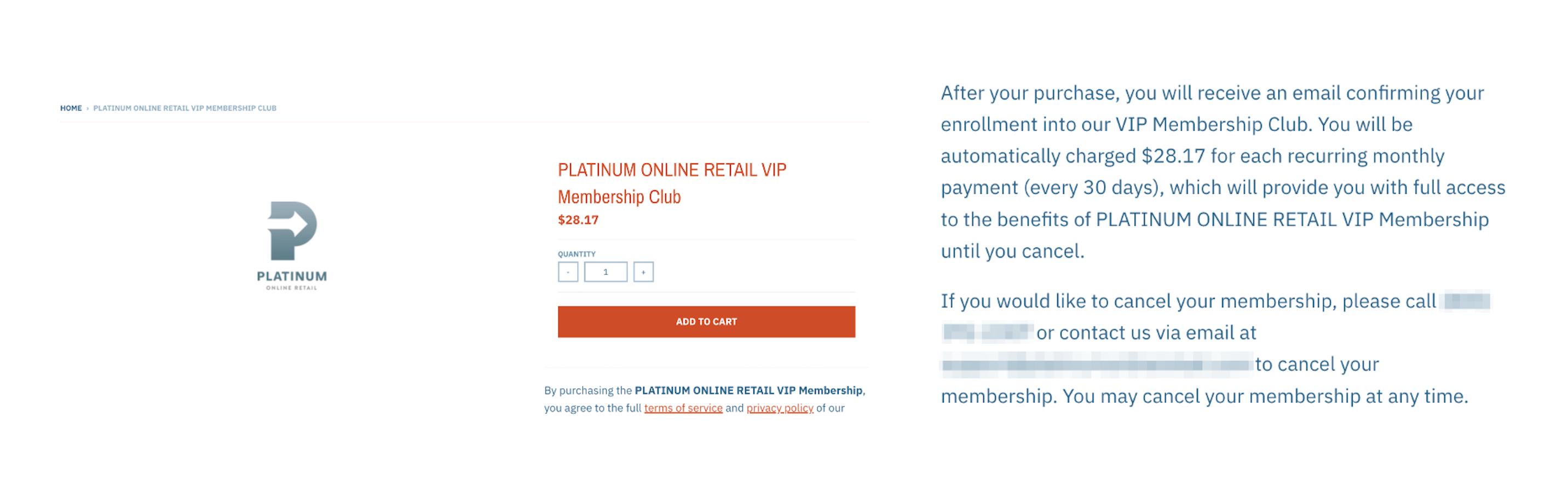}
\caption{An Example of Dark Patterns (platinumonlineretail.com)}
\label{dpex}
\end{figure*}

Prior studies \cite{dpYoutube,11kScale,mobileApp,cookieConsent} have reported that dark patterns exist everywhere on online services, including e-commerce sites, social networking services (SNS), consent to cookies, and apps. 

\subsection{Dark Patterns and privacy}

Dark patterns also pose problems to user privacy protection, including UX designs that induce users to provide personal data or consent to cookies in online services. Discussions on the impact of dark patterns to protect user privacy are not limited to academic research and have been widely discussed in various places.

In 2018, the California Legislature passed the Consumer Privacy Protection Act (CCPA) \cite{ccpa} to ban dark patterns on the Internet, which became effective in 2020 and had a critical impact on privacy-related choices. In 2019, the Commission Nationale de l'Informatique et des Libertés (CNIL) in France published a report \cite{cnil} on the impact of UX design on privacy protection. The report argued that manipulative and/or misleading interfaces on online services could influence critical decisions related to user privacy. The report also raised awareness of such dark patterns and called on designers to collaborate for privacy-friendly designs. In 2020, the Organization for Economic Development and Cooperation (OECD) discussed the privacy and purchasing behavior risks that dark patterns pose to consumers \cite{oecd}. During the meeting, the risk of dark patterns exposing personal information on online services without the consumer’s genuine consent was mentioned. In 2021, the European Data Protection Board (EDPB) discussed dark patterns in social media that can negatively impact users’ decisions regarding the handling of personal information \cite{edpb}. The main objective was to discuss protecting users from dark patterns that may harm their privacy.

As ever-increasing dark patterns have become a social problem, as evidenced by the policies of various countries, urgent actions are required.

\subsection{Study Contributions} 

Prior works \cite{dpYoutube,11kScale,mobileApp,cookieConsent,darkSideUx,darkpattern.org} investigated and classified dark patterns on e-commerce sites, social networking services sites, news sites, apps, and cookie consent. However, to the best of our knowledge, auto-detection of dark patterns was outside their scope. Although a few studies \cite{webbasedDp,autoCookie} tackled auto-detection, they relied on manual methods to extract features for auto-detection.
 
 This work distributed a dataset of dark patterns consisting of positive and negative samples, which will help new research in this area to detect dark patterns automatically on websites by distributing a dataset of dark patterns consisting of positive and negative samples. Besides, we provide baseline results of auto-detection with state-of-the-art machine learning methods, including BERT \cite{bert}, RoBERTa \cite{roberta}, ALBERT \cite{albert}, and XLNet \cite{xlnet}, for easy comparison with future algorithms. 

\subsection{Organization of the Paper}

Related work is introduced in Section \ref{sec:relatedWork}, followed by a description of the creation of the dark pattern dataset in Section \ref{sec:ecommerceDarkPatternDataset}. Section \ref{sec:baselineMethodsForAutomaticDetection} investigates the performance of auto-detection with state-of-the-art machine learning algorithms as baselines. Finally, we conclude this paper in Section \ref{sec:conclusion}.

% Our contributions can be summarized as:
% \begin{itemize}
%     \item We constructed a text-based dataset for dark patterns automatic detection on online shopping sites.
%     \item We provided baseline performances of text-based dark patterns automatic detection using classical NLP methods and transformer-based pre-trained language models.
%     \item We experimentally show that high performance text-based dark patterns automatic detection is possible by applying text classification techniques known as state-of-the-arts.
% \end{itemize}

\section{Related Work}\label{sec:relatedWork}

This section introduces related work from three categories: dark pattern taxonomies, dark patterns at scale, and automated dark pattern detection.

\subsection{Dark Pattern Taxonomies}

The taxonomies of dark patterns have been defined in various ways, including the research aiming to define the taxonomies of dark patterns and the research classifying dark patterns from the gathered large-scale web pages.

The term “dark patterns” was first defined on Harry's website \cite{darkpattern.org} in 2010. He classified dark patterns into 12 types and introduced examples of each type. In 2018, Gray et al. \cite{darkSideUx} gathered examples of dark patterns through searches with the keyword “dark patterns” and its derivatives on Google, Bing, and social networking sites, such as Facebook and Instagram. Gray et al. analyzed the contents to classify dark patterns into 5 types. Table \ref{tab:dptype:graym} lists the types of dark patterns defined by Gray et al. In 2021, Mathur et al.\cite{whatDp} conducted a comprehensive study to summarize the various definitions of dark patterns, which included 84 types of dark patterns defined by 11 studies in the fields of human-computer interaction, security and privacy, law, and psychology.
\begin{table*}[t]
	\centering
	\caption{Types of Dark Patterns Defined by Gray et al. \cite{darkSideUx}}
	\def\arraystretch{1.5}%  1 is the default, change whatever you need
	\begin{tabularx}{\textwidth}{l|X}  \hline
Types of Dark Patterns & Description \\ \hline
  Nagging & Redirection of expected functionality that persists beyond one or more interactions. \\ 
Obstruction & Making a process more difficult than it needs to be, with the intent of dissuading certain action(s) \\ 
Sneaking & Attempting to hide, disguise, or delay the divulging of information that is relevant to the user. \\
Interface Interference & Manipulation of the user interface that privileges certain actions over others. \\ 
Forced Action & Requiring the user to perform a certain action to access (or continue to access) certain functionality \\ \hline
	\end{tabularx}
	\label{tab:dptype:graym}
\end{table*}

As described above, the taxonomies of dark patterns have been defined based on various criteria in several studies.

\subsection{Dark Patterns at Scale}

Many researchers have conducted studies on dark patterns at a large scale across online content to determine the extent to which dark patterns exist on online services.

In 2018, Mathur et al. \cite{dpYoutube} conducted a large-scale study to determine whether influencers disclose that their content is advertising when they promote specific products. The Federal Trade Commission (FTC) prohibits creators from promoting specific products without disclosing the fact to users. Mathur et al. found that approximately 10\% of the affiliate advertisements content did not disclose that they were advertisements. In 2019, Mathur et al.\cite{11kScale} extensively studied dark patterns on shopping websites. They obtained 11,286 shopping sites by extracting English shopping sites from 361,102 websites with the highest number of accesses obtained by Alexa Traffic Rank. As a result, they found 7 types of 1,818 dark patterns from 1,254 shopping sites, which is approximately 11.1\% of the total, and they published the dark pattern text data \footnote{https://github.com/aruneshmathur/dark-patterns}.

In 2020, Di Geronimo et al. \cite{mobileApp} studied 240 popular Android applications. Two researchers performed a set of common activities, including user registration and configuration changes, to investigate whether dark patterns existed in the applications. Di Geronimo et al. analyzed the dark patterns found and classified them into the five types of dark patterns defined by Gray et al. \cite{darkSideUx}.

In 2020, Soe et al. \cite{cookieConsent} conducted a large-scale study of dark patterns in cookie consent notifications. The work targeted the consent notices for cookies on 300 news sites written in Nordic and English. Two researchers asked the following questions: “Are there dark patterns in the consent notices?”, If so, “what is the type of dark pattern (of the five types of dark patterns defined by Gray et al.)?”, “Is it possible to refuse cookies?”, and “Where does the cookie consent notice appear?”. As a result, they identified 297 dark patterns for inducing consent to cookies.

\subsection{Automated Detection for Dark Patterns}

A few studies have addressed the automatic detection of dark patterns as follows.

In 2021, Andrea et al. \cite{webbasedDp} proposed a detection framework for dark patterns with a combination of manual and automatic methods. They targeted the 12 dark pattern types defined by Harry \cite{darkpattern.org}. The weakness of the framework is that only simple keyword matching is adopted for automated detection methods. For example, the automated detection is based on whether the keywords opt-in or opt-out are included or not. Other features to detect dark patterns rely on manual methods, so the framework cannot detect dark patterns automatically.

In 2022, Soe et al. \cite{autoCookie} conducted automatic dark pattern detection in cookie banners. They experimented using machine learning techniques on a manually collected dataset of cookie consents on 300 news websites \cite{cookieConsent}. They used manually extracted features, such as text in banners, location of the pop-up, number of clicks to reject all consent, purpose of the cookies' inclusion, and the existence of any third-party cookies. The automatic detection targeted categorizing 5 types of dark and non-dark patterns by adopting 10 features, where the 5 types of dark patterns were those defined by Gray et al. [7]. The experimental results with gradient-boosted tree classifiers showed accuracies of 0.50 (obstruction) to 0.72 (nagging). The weaknesses of this research are 1) only targeting cookie banners, 2) features were extracted manually, and 3) the obtained accuracies were low.

\section{E-commerce Dark Pattern Dataset}\label{sec:ecommerceDarkPatternDataset}

The purpose of the proposed dataset is to enable a wide range of research for automatic dark pattern detection on e-commerce sites. Although we may prepare various features that Soe et al. \cite{autoCookie} adopted, we only prepare texts that could be automatically extracted from web pages because we target automatic detection without manually extracted features. Our dataset is inspired by Mathur et al.’s work in 2019 \cite{11kScale}, which consisted of 1,818 dark pattern texts from shopping sites. We added non-dark pattern texts to Mathur et al.’s dataset to prepare a dataset with a balanced number of dark and non-dark pattern texts. 

\subsection{Dark Pattern Texts in E-commerce Sites}\label{sbsec:dpTextInEcomSites}

We modified the dataset of dark patterns manually constructed by Mathur et al. \cite{11kScale}, where 1,818 dark pattern texts from 1,254 shopping sites, approximately 11.1\% of the 11,000 shopping sites they studied, were included. The original Mathur et al. dataset consists of the manually tagged features listed in Table \ref{tab:feature:mathur}. As our goal is to prepare a dark pattern text dataset, we excluded missing, i.e., null in the “Pattern String” field in Table \ref{tab:feature:mathur}, or duplicate text data from the original dataset, followed by tagging them as dark patterns, i.e., positive examples. Finally, we have 1,178 text data.

\begin{table}[b]
	\centering
\caption{Mathur et al.’s Dataset Features\cite{11kScale}}
	\def\arraystretch{1.5}%  1 is the default, change whatever you need
	\begin{tabularx}{250px}{l|X}   \hline
		Feature Name      & Description                                   \\ \hline
		Pattern String    & Text of dark pattern                          \\ 
		Comment           & Comments from researchers                     \\
		Pattern Category  & Category of dark patterns
                   \\
		Pattern Type      & Detailed type of dark patterns (e.g., trick question, hard to cancel) \\
		Where on the website? & Where on the website is the dark pattern present \\ 
		Deceptive?        & Whether the dark pattern instance is deceptive or not \\ 
		Website Page      & Page's URL that has dark patterns       \\ \hline 
	\end{tabularx}
	\label{tab:feature:mathur}
\end{table}

\subsection{Non-Dark Pattern Texts in E-commerce Site}\label{sbsec:nonDpTextInEcomSites}
The collection of negative samples, i.e., non-dark pattern texts, is described in this section. The following two steps were conducted to create non-dark pattern texts:

\begin{enumerate}
   \item Collect the web pages on e-commerce sites.
   \item Extract texts from the collected web pages to segment.
   \item Exclude dark patterns from the segmented texts.
\end{enumerate}

\subsubsection{Collecting web pages} 

The negative samples were retrieved from the same websites, i.e., e-commerce sites, where the dark patterns prepared in Section \ref{sbsec:dpTextInEcomSites} are included, using headless Chrome. If a website was unreachable, encountering “not found” or “access denied,” we ignored the website. The content was gathered after executing JavaScript when accessing each web page because most websites adopt JavaScript to create the web page.

% TODO:
\begin{algorithm*}[t]
    \newcommand{\Continue}{\textbf{continue;}}
    \newcommand{\End}{\textbf{end}}
    \label{alg:segmentation}
    \caption{Segmentation Algorithm  (modified from Algorithm 1 of Mathur et al. \cite{11kScale})}
    \SetKwInOut{Input}{Input}
    \SetKwInOut{Output}{Output}
    \LinesNumbered
    \Input{$element$: HTML Element \\ $ignoreElements$: Tag names ignored in segmentation ['script','style','noscript','br','hr']   \\
    $blockElements$: Tag names('p','div',...) of all Block-level Elements \\
    $inlineElements$: Tag names('button','span',...) of all Inline Elements  \\
    }
    \Output{text list of segmented html $texts$ \\}
    % \SegmentElement
    \textbf{function segmentElement(element):} \\
    \Begin{
       \If{$element$ is null}
        {\KwRet {[empty list]}}
     
     \If{All child nodes of $element$ are not included in $ignoreElements$}
        {

        \If{All child nodes of $element$ are TEXT\_NODE or included in $inlineElements$}
            {
                $text \leftarrow$ text content of $element$ \;
                \If{$text$ is not null}
                    {
                        \KwRet{[$text$]}
                    }
            }
        }

     $childNodes \leftarrow$ child nodes of $element$ \;
     $texts \leftarrow$ [empty list] \;
     \For {$i = 0$ to $|childNodes|$}{
        $child \leftarrow childNodes[i]$\;
        $nodeType \leftarrow $ node type of $child$\;
        \If{$nodeType =$ TEXT\_NODE}
            {
                $textContent \leftarrow$ text content of $child$ \;
                \If{$textContent \ne$  null}{
                    Append $textContent$ to $texts$ \;
                }
            }
        $tagName \leftarrow $ tag name of $child$ \;
        \If{($tagName = $ null) OR ($tagName \in ignoreElements$)}{
            \Continue
        }

        \If{$tagName \in blockElements$}
            {
                $childTexts \leftarrow segmentElements(child)$ \;
                Concatenate $texts$ with $childTexts$ \;
            }
        
        \If{$tagName$ is included in $inlineElements$}{
            $textContent \leftarrow$ text content of $child$ \;
            \If{$textContent \ne$ null}{
                Append $textContent$ to $texts$
            }
        }
     }
     \KwRet $texts$
    }
    \End
\end{algorithm*}

\subsubsection{Extracting texts}

After collecting web pages, we adopted a Puppeteer\footnote{\url{https://github.com/puppeteer/puppeteer}} library to scrape each content, followed by collecting its screenshot and text. Mathur et al. targeted the text inside the UI components, not the whole web page, as the unit of extraction for dark patterns. Besides, Mathur et al.'s \cite{11kScale} restricts the target HTML element to be 1) a UI element that occupies more than 30\% of the page and 2) a specific block element to obtain dark pattern candidates efficiently. On the contrary, our goal is to extract non-dark pattern texts widely from the page so that we target all the block elements to extract the text. Specifically, we target the block elements containing at least one TextNode as their child element but not containing block-level elements \footnote{\url{https://developer.mozilla.org/en-US/docs/Web/HTML/Block-level\_elements}}. Because the only difference from the work of Mathur et al. is target elements, we can use the same segmented texts. The detailed segmentation algorithm is given as Algorithm \ref{alg:segmentation}, which is modified from Mathur et al.'s algorithm. We applied Algorithm \ref{alg:segmentation} to the body element of each web page.

\subsubsection{Excluding dark patterns} 

The collected texts in the previous step may contain dark patterns, so we filtered out the texts collected by Mathur et al. that contain dark pattern texts. Then, we manually confirmed the filtered texts were non-dark patterns. Note that we ignored numerical values, capitalization, and punctuation in the text when applying the filtering.

Through the above process, 14,208 non-dark pattern texts were collected. From these, 1,178 texts were extracted randomly and used as negative samples of the dataset.

\subsection{Validation of the Segmentation Algorithm} 

To validate the segmentation by the implemented Algorithm \ref{alg:segmentation}, we segmented the same web pages included in Mathur et al.'s dataset into texts. Then, we manually checked whether the segmented texts contained the same units of dark pattern texts. Table \ref{tab:dp:cmp} shows a comparison of the dark patterns collected by Mathur et al. and an example of the text obtained by Algorithm  \ref{alg:segmentation}. As shown in Table \ref{tab:dp:cmp}, we confirmed the texts are the same as Mathur et al.'s dark pattern texts except for numerical values and punctuations.
\begin{table*}[t]
	\centering
	\caption{Comparison of Text Extraction Units}
	\def\arraystretch{1.5}
	\begin{tabularx}{\textwidth}{X|X|X}  \hline
 Websites URL & Dark patterns texts by Mathur et al. & Texts by our segmentation algorithm \\ \hline
\url{anuradhaartjewellery.com} & "Hurry Up! Only 1 Piece Left" & "Hurry Up! Only 1 Piece Left" \\ 
\url{annthegran.com} & "No thanks, I don't like free stuff" & "No thanks, I don't like free stuff" \\ 
\url{savethebeesproject.com} & "9 people are viewing this." &  "24 People are viewing this!" \\ \hline
\end{tabularx}
\label{tab:dp:cmp}
\end{table*}

For illustrative purposes, we show several examples of the segmentation results in Table \ref{tab:segment:result}.

\begin{table*}[t]
        \centering
	\caption{Examples of HTML Segmentation by Algorithm \ref{alg:segmentation}}
	\def\arraystretch{1.5}%  1 is the default, change whatever you need
	\begin{tabular}{l|l}  \hline
            $element$(Input) & $texts$(Output) \\ \hline
            \begin{lstlisting}
<body>
    <p>This is in p (Block-level) tag. </p>
</body>
\end{lstlisting} & [
      'This is in p (Block-level) tag. '
    ] \\ \hline
  \begin{lstlisting}
<body>
    <div>
        <p>This is in p (Block-level) tag. </p>
        <p>This is in p (Block-level) tag. </p>
    </div>
</body>
\end{lstlisting} & [ 'This is in p (Block-level) tag. ', 'This is in p (Block-level) tag. ' ]
 \\ \hline
  \begin{lstlisting}
<body>
    <div>
        <p>This is in p (Block-level) tag. 
            <span>This is in span (Inline-level) tag. </span>
        </p>
    </div>
</body>
\end{lstlisting} & [
      'This is in p (Block-level) tag. This is in span (Inline-level) tag. '
    ] \\ \hline
  \begin{lstlisting}
<body>
    <p>This is in p (Block-level) tag. </p>
    <script>console.log("script will be ignored")</script>
</body>
\end{lstlisting} & [
      'This is in p (Block-level) tag. '
    ] \\ \hline
	\end{tabular}
	\label{tab:segment:result}
\end{table*}

\subsection{Summary}
The final dataset consists of 1,178 positive (dark pattern) and 1,178 negative (non-dark pattern) texts, totaling 2,356 texts from e-commerce websites. 

An example of the data is shown in Table \ref{tab:dp:example}. In this example, the dark pattern taxonomy of the first positive example "3,081 people have viewed this item" is Social Proof, which misleads users to purchase by displaying information as if other users have purchased the product. It exploits the bandwagon effect \cite{11kScale}. We published the dataset on Github\footnote{https://github.com/yamanalab/ec-darkpattern} with the code used to collect non-dark pattern texts.

\begin{table*}[t]
	\centering
	\caption{Examples of Positive and Negative Texts}
	\def\arraystretch{1.5}
	\begin{tabularx}{\textwidth}{X|X}  \hline
	Positive (dark pattern) & Negative (non-dark pattern)  \\ \hline
    "3,081 people have viewed this item" & "Unique and personalized products to go as yourself." \\ 
    "In Stock only 3 left" & "newsletter signup (privacy policy)." \\
       "No thanks. I don't like free things..." & "International Shipping Policy" \\ 
       "894 Claimed! Hurry, only a few left!" & "Clothes, Shoes \& Accessories" \\ 
       "Your order is reserved for 08:48 minutes!" & "READY FOR YOUR NEXT AUTHENTIC NHL JERSEY?" \\ 
    \hline
	\end{tabularx}
	\label{tab:dp:example}
\end{table*}

% 出力されるレイアウトの都合でここに配置する。

\section{Baseline Methods for Automatic Detection}\label{sec:baselineMethodsForAutomaticDetection}

\subsection{Selection of Baseline Methods}

In recent years, many machine learning-based text classification methods have been proposed. In the security and privacy field, fake news and phishing detection are representative text classification tasks. Various machine learning-based methods have been proposed for text-based fake news and phishing detection, including classical machine learning and deep learning-based methods.

In natural language processing, deep learning-based models have surpassed the accuracy of classical machine learning models in various text classification tasks. Among deep learning-based models, models adopting transformer-based pre-trained language models like BERT \cite{bert} have attracted the most attention and are the current state-of-the-art.

Thus, we chose the following two types of machine learning methods as the baseline methods:

\begin{itemize}
    \item Classical NLP methods: logistic regression, SVM, random forest, and gradient boosting (LightGBM \cite{lgbm}).
    \item Transformer-based pre-trained language models: BERT \cite{bert}, RoBERTa \cite{roberta}, ALBERT \cite{albert}, and XLNet \cite{xlnet}.
\end{itemize}

\subsection{Metrics}

Automatic detection for dark patterns is a binary classification that predicts whether a given text is a dark or non-dark pattern. We adopted accuracy, precision, recall, F1-score, and AUC to evaluate the models over 5-fold cross-validation.

\subsection{Experimental Evaluations using Classical NLP Methods}

We adopted the following two-step procedure for classical NLP methods:

\begin{enumerate}[label=(\roman*)]
    \item Extract features of the texts using bag-of-words (BoW), followed by feeding each classifier.
    \item Train each dark pattern classifier. 
\end{enumerate}

We used scikit-learn\footnote{https://github.com/scikit-learn/scikit-learn} to implement the logistic regression, SVM, and random forest classifiers. For gradient boosting (LightGBM), we used lightgbm\footnote{https://github.com/microsoft/LightGBM}. We tuned hyper-parameters by using Optuna \cite{optuna}. Table \ref{tab:classicalNlp:hyper} in the Appendix shows the best hyper-parameters for the classical NLP methods  we adopted.

The experimental evaluation results are listed in Table \ref{tab:classicalNlp:result}, showing accuracies of 0.954 to 0.962.

\begin{table}[b]
	\centering
	\caption{Experimental Result of Classical NLP Methods}
	\def\arraystretch{1.5}
	\begin{tabular}{cccccc}  \hline
		Model & Accuracy & AUC &F1 score  & Precision & Recall \\ \hline
		Logistic Regression & 0.961 & 0.989 & 0.960  & 0.981 & 0.940 \\ 
		SVM & 0.954 & 0.987 & 0.952  & 0.986 & 0.922 \\ 
		Random Forest & 0.958 & 0.989 & 0.957 & 0.984 & 0.932 \\ 
		Gradient Boosting & 0.962 & 0.989 & 0.961 & 0.976 & 0.947 \\ \hline	
	\end{tabular}
	\label{tab:classicalNlp:result}
\end{table}

\subsection{Experimental Evaluations using Transformer-based Pre-trained Language Models }

BERT is a model with multiple layers of transformer encoders. BERT acquires knowledge about languages and domains through pre-training with a masked language model (MLM) and next sentence prediction (NSP) using a large-scale text corpus.

When applying BERT to text classification, it is common to perform transfer learning and fine-tuning on a task-specific dataset based on BERT that has been pre-trained on the language corpus targeted by the task. An overview of the BERT-based model for automatic dark pattern detection used in this study is shown in Fig. \ref{bertfig}. "[CLS]" and "[SEP]" are special tokens. "[CLS]" is used for classification as sentence representation. "[SEP]" is used for separating segments. The model was trained so that the classification result is output from the classification layer that linearly transforms the "[CLS]" representation output by BERT. 

\begin{figure}[t]
\includegraphics[width=\linewidth]{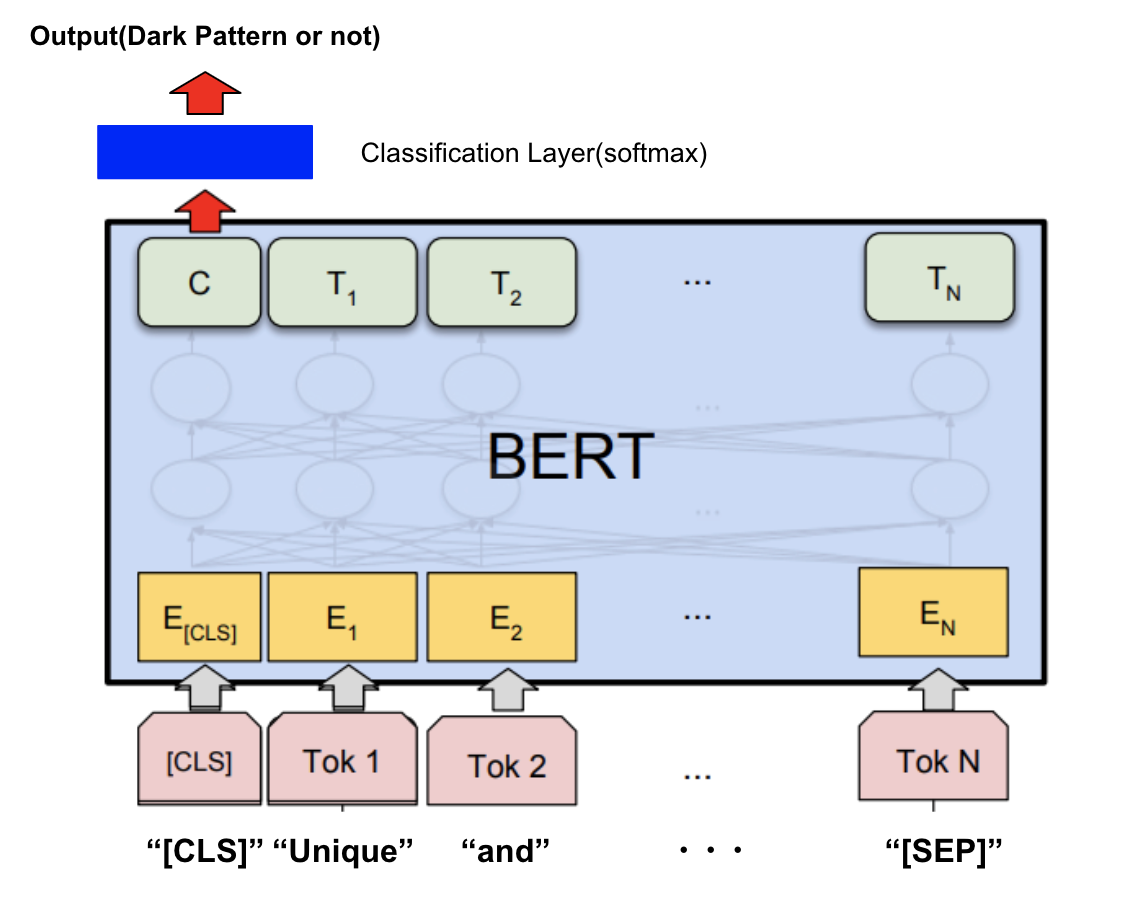}
\caption{Overview of Transformer-based Pre-trained Language Models}
\label{bertfig}
\end{figure}

We used the transformers\footnote{https://huggingface.co/transformers/} to develop and train the BERT-based models. We used AdamW \cite{adamw} optimizer and linear learning-rate scheduler. When training deep learning models, we tuned hyper-parameters by grid search. Table \ref{tab:dlhyper:result} in the Appendix lists the best hyper-parameters for the transformer-based pre-trained language models we adopted.

The experimental results of transformer-based pre-trained language models are listed in Table \ref{tab:dl:result}, which shows the best accuracy of 0.975 using $\text{RoBERTa}_{large}$.

\begin{table}[b]
	\centering
	\caption{Experimental Results of Transformer-based Pre-trained Language Models}
	\def\arraystretch{1.5}
	\begin{tabular}{cccccc}  \hline
		Model  & Accuracy & AUC &F1 score  & Precision & Recall \\ \hline
		$\text{BERT}_{base}$ & 0.972 & 0.993 & 0.971 & 0.982 & 0.961  \\ 
		$\text{BERT}_{large}$ & 0.965 & 0.993 & 0.965 & 0.973 & 0.957  \\
		$\text{RoBERTa}_{base}$ & 0.966	& 0.993	& 0.966	& 0.979	& 0.954  \\ 
		$\text{RoBERTa}_{large}$ & $\mathbf{0.975}$ & $\mathbf{0.995}$ & $\mathbf{0.975}$ & $\mathbf{0.984}$ & $\mathbf{0.967}$  \\ 
		$\text{ALBERT}_{base}$ & 0.959 & 0.991 & 0.959 & 0.972 & 0.946  \\ 
		$\text{ALBERT}_{large}$ & 0.965 & 0.986 & 0.965 & 0.973 & 0.957  \\ 
		$\text{XLNet}_{base}$ & 0.966 & 0.992 & 0.966 & 0.975 & 0.958  \\ 
		$\text{XLNet}_{large}$ & 0.942 & 0.988 & 0.940 & 0.968 & 0.914  \\ \hline
	\end{tabular}
	\label{tab:dl:result}
\end{table}

\section{Conclusion}\label{sec:conclusion}

This study constructed a dark pattern dataset for e-commerce sites with baseline automatic detection performance. We hope the dataset will help researchers to pursue various research on the automatic detection of dark patterns. As for the baseline performance, we experimented with automatic dark pattern detection using a set of machine learning methods, including classical NLP-based models and transformer-based pre-trained language models. The results show that the RoBERTa-large model achieved a maximum accuracy of 0.975.
 
Our future work will include 1) enhancing the dataset to include other websites, not just e-commerce sites, and 2) expanding the dataset to include other UX-related features besides texts, such as images and scripts because UX-related features are also helpful in detecting dark patterns.

\bibliographystyle{IEEEtran} %引用された順番に出力
\bibliography{darkpattern} %bibファイルの.bibの前の部分

\newpage

\appendix

\begin{table}[h]
        \centering
	\caption{Hyper-parameters of Classical NLP Methods}
	\def\arraystretch{1.5}
	\begin{tabular}{ccc}  \hline
	Model & Parameter name & Parameter value  \\ \hline
        Logistic Regression 
            & C & 4.95 \\  \hline	
          \multirow{2}{*}{SVM} 
            & kernel & rbf \\
            & C & 4.35 \\  \hline	
          \multirow{5}{*}{RandomForest}  
            & max\_depth & 32 \\ 
	    & n\_estimators & 641 \\ 
		& min\_samples\_split & 18 \\ 
		& min\_samples\_leaf & 1 \\ 
		& bootstrap & False \\  \hline	
         \multirow{7}{*}{LightGBM}   & lambda\_l1 & $3.04 \times 10^{-8}$ \\ 
		  & lambda\_l2 &  0.806 \\ 
		& num\_leaves & 89 \\ 
            & feature\_fraction & 0.466 \\ 
            &	bagging\_fraction & 0.887 \\ 
            &	bagging\_freq & 6 \\ 
            &	min\_child\_samples & 5 \\ \hline
	\end{tabular}
	\label{tab:classicalNlp:hyper}
\end{table}

\begin{table}[h]
        \centering
	\caption{Hyper-parameters of Transformer-based Pre-trained Language Models}
	\def\arraystretch{1.5}
	\begin{tabular}{ccccc}  \hline
		Model  & Batch Size & Learning Rate & Dropout rate  & Epochs \\ \hline
		$\text{BERT}_{base}$ & 16 & 4e-5 & \multirow{8}{*}{0.1} & \multirow{8}{*}{5}  \\
		$\text{BERT}_{large}$ & 32 & 3e-5 &  &   \\ 
		$\text{RoBERTa}_{base}$ & 128	& 3e-5	& 	&   \\ 
		$\text{RoBERTa}_{large}$ & 32 & 3e-5 &  &   \\ 
		$\text{ALBERT}_{base}$ & 16 & 3e-5 &  &   \\ 
		$\text{ALBERT}_{large}$ & 32 & 5e-5 &  &  \\ 
		$\text{XLNet}_{base}$ & 16 & 2e-5 &  &  \\ 
		$\text{XLNet}_{large}$ & 32 & 4e-5 &  &   \\ \hline
	\end{tabular}
	\label{tab:dlhyper:result}
\end{table}

\end{document}